\documentclass{article}
\usepackage{spconf,amsmath,graphicx}
\usepackage{enumitem}
\setlist{nosep, leftmargin=14pt}
\usepackage{mwe} 

\usepackage{amsfonts}
\usepackage{booktabs}
\usepackage{colortbl}
\usepackage[table]{xcolor} 
\definecolor{lightblue}{RGB}{173,216,230}

\title{Where to Begin? From Random to Foundation Model Instructed Initialization in Federated Learning for Medical Image Segmentation}

\name{Ming Li$^{1,2}$, Guang Yang$^{1,2,3,4}$\thanks{Correspondence: \texttt{g.yang@imperial.ac.uk}}}
\address{
$^{1}$Bioengineering Department and Imperial-X, Imperial College London, London W12 7SL, UK.\\
$^{2}$National Heart and Lung Institute, Imperial College London, London SW7 2AZ, UK.\\
$^{3}$Cardiovascular Research Centre, Royal Brompton Hospital, London SW3 6NP, UK.\\
$^{4}$School of Biomedical Engineering \& Imaging Sciences, King's College London, London WC2R 2LS, UK.
}

\begin{document}

\sloppy
\maketitle

\begin{abstract}
In medical image analysis, Federated Learning (FL) stands out as a key technology that enables privacy-preserved, decentralized data processing, crucial for handling sensitive medical data.
Currently, most FL models employ random initialization, which has been proven effective in various instances. However, given the unique challenges posed by non-IID (independently and identically distributed) data in FL, we propose a novel perspective: exploring the impact of using the foundation model with enormous pre-trained knowledge, such as the Segment Anything Model (SAM), as an instructive teacher for FL model initialization in medical image segmentation task. 
This work for the first time attempts to utilize the foundation model as an instructive teacher for initialization in FL, assessing its impact on the performance of FL models, especially in non-IID data scenarios.
Our empirical evaluation on chest x-ray lung segmentation showcases that FL with foundation model instructed initialization not only achieves faster convergence but also improves performance in complex data contexts. 
These findings offer a new perspective for model initialization in FL.

\end{abstract}

\begin{keywords}
Federated Learning, Foundation Model, Initialization
\end{keywords}

\section{Introduction}

\begin{figure}[!th]
    \centering
    \includegraphics[width=0.45\textwidth]{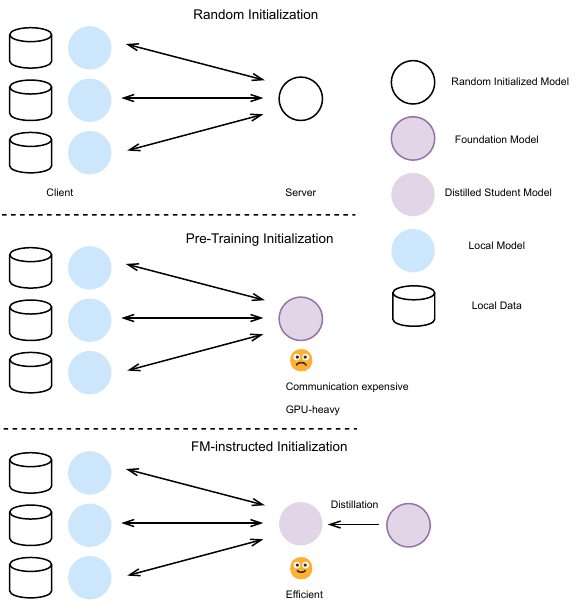}
    \caption{Where to begin? From random initialization to foundation model instructed initialization. Considering the high communication costs and GPU resource demands of directly employing the foundation model, it is used as an instructive teacher for the initialization in FL.}
\end{figure}

Medical image analysis, particularly segmentation, is a critical area in healthcare, aiding in the accurate diagnosis and treatment of various diseases.
Despite its importance, this task faces significant challenges in terms of data privacy and the heterogeneity of data sources
\cite{liu2021deep,li2020mv,li2019recurrent}.

Federated Learning (FL) \cite{mcmahan2017communication} has emerged as a key technology in the context of data privacy and decentralization. It allows for collaborative model training across multiple data sources without the need to share raw data, which is crucial in handling sensitive medical data. 
Typically, FL begins model training from a random initialization. While random initialization is effective in various scenarios, it presents challenges, notably in handling non-IID (independently and identically distributed) data, a common scenario in medical datasets where data distribution varies significantly across different sources \cite{li2023data}.

Pre-training is commonly applied in computer vision, natural language processing, and many other application domains to speed up convergence and boost accuracy for downstream tasks \cite{dong2018holistic,li2020unified}. 
Despite the ample research on pre-training, its impacts on FL model initialization for medical image analysis remain largely unexplored \cite{julka2023knowledge,wu2023segment}. 
The concept of Foundation Model (FM) \cite{bommasani2021opportunities}, such as the Segment Anything Model (SAM) \cite{kirillov2023segany}, offers a new perspective. These models are pre-trained on vast datasets and carry extensive knowledge, which can be beneficial in addressing the non-IID challenge in FL \cite{chen2023importance,nguyen2023begin}. 
This work explores the potential of using FM, specifically SAM-Med2D \cite{cheng2023sam}, a variant of SAM fine-tuned for medical imaging, as an instructive teacher for FL model initialization. The objective is to assess whether the pre-trained knowledge of FM can help mitigate the non-IID dilemma in FL, thus enhancing model performance and efficiency.

However, the direct application of large FM like SAM-Med2D in FL poses significant challenges, primarily due to its size and consequent resource intensiveness. FM is GPU-heavy and can lead to substantial communication costs during the FL process, posing a significant hurdle for efficient FL deployment. To address this issue, we propose the usage of knowledge distillation technique \cite{hinton2015distilling}. The idea is to distill the knowledge from the large FM into a smaller, more manageable model, which can then serve as the initialization model in FL. This approach aims to leverage the strengths of FM while mitigating its resource and communication demands.

In summary, this work aims to investigate the impact of using FM, particularly SAM-Med2D, as an instructive teacher for model initialization in FL within the domain of medical image segmentation. 
Our study on chest x-ray lung segmentation reveals three findings:
\begin{itemize}
    \item FM instructed initialization can serve as a good starting point for FL, enabling models to converge faster and achieve better performance, without numerous communication rounds, thus avoiding huge communication cost.
    \item Utilizing knowledge distillation, where the FM acts as the ``teacher," enhances the performance and generalization capabilities of simpler ``student" model within the FL system. 
    \item Starting from FM-instructed initialization also mitigates the impact of non-IID data.
\end{itemize}


\section{Method}
\subsection{Federated Learning}
In the context of FL, the training data are distributed among $K$ clients.
Each client $k$ has a local private dataset $D^{k} = \{(x_{i}, y_{i})\}^{|D^{k}|}_{i=1}$, where $|D^{k}|$ is the number of data samples within client $k$, $x_{i}$ is the data sample, and $y_{i}$ is the groundtruth.
FL aims to learn a global model parameterized by $\theta$ that solves the following optimization problem and minimizes its risk on all clients:
\begin{equation}
\begin{split}
    \min _{\theta} \mathcal{L}(\theta) = \sum_{k=1}^K \frac{|D^{k}|}{|D|} \mathcal{L}_{k}(\theta), \\
    \text{where} \quad \mathcal{L}_k(\theta) = \frac{1}{|D_k|} \sum_{i=1}^{|D_k|} \ell(x_i, y_i; \theta).
\end{split}
\end{equation}
Here, $D$ represents the training data of all clients, $\ell$ is the loss function on a single data sample, $\mathcal{L}_k$ signifies the empirical risk on client $k$, and $\mathcal{L}$ denotes the empirical risk on $D$.
\par

Federated Averaging (FedAvg) \cite{mcmahan2017communication}, the most widely used and standard method in FL, consists of alternating between concurrent multiple local stochastic gradient descent update at each client and the model aggregation update at the central server over several communication rounds.
The local client update and global server aggregation in FL can be defined as:
\begin{align}
    &\text{Client:} \quad \theta_k^{(t)} = \arg\min_{\theta} \mathcal{L}_k (\theta), \; \text{initialized by } \theta^{(t-1)}, \\
    &\text{Server:} \quad \theta^{(t)} = \sum_{k=1}^{K} \frac{|D_k|}{|\mathcal{D}|} \theta_k^{(t)}.
\end{align}
Here, $\theta_k^{(t)}$ represents the local model parameters of client $k$ after the $t$-th communication round, $\theta^{(t)}$ denotes the global model parameters after the $t$-th round of communication. The objective in the local training phase is to minimize the empirical risk per client, often employing several epochs of stochastic gradient descent updates. Subsequently, the global aggregation phase computes an element-wise average of all local models.
\par

Vanilla FedAvg, starting from a random initialization, demonstrates a vulnerability to the heterogeneity in client data \cite{chen2023importance,nguyen2023begin}. 
The non-IID nature of client data can lead to significant divergence in the local models both from each other and from the global optimum.
Such divergence often results in a clear decline in the model performance. 
In this study, we employ FedAvg to explore the influence of different initialization strategies in FL, aiming to understand how various initial conditions impact the convergence and effectiveness of FedAvg in handling non-IID data heterogeneity, with a specific focus on the medical image segmentation task.

\subsection{Knowledge Distillation for FM}
Knowledge distillation \cite{hinton2015distilling} is an effective technique for transferring knowledge from a larger, well-trained teacher model to a smaller and efficient student model. This approach is particularly valuable in scenarios where deploying a large model is computationally prohibitive.
\par

SAM-Med2D \cite{cheng2023sam}, a foundational model for segmenting medical 2D images, exemplifies the power of large-scale training. It has been trained on a massive dataset of 4.6 million images and 19.7 million masks, endowing it with exceptional zero-shot generalization capabilities. However, the sheer size of SAM-Med2D makes it impractical for direct use in FL, which often requires lightweight models due to computational and bandwidth constraints.
\par

Inspired by \cite{julka2023knowledge,wu2023segment}, we employ knowledge distillation to transfer the rich insights from SAM-Med2D to a more compact model, SM-lite. This process involves using the outputs of SAM-Med2D to guide the learning process of SM-lite. We freeze SAM-Med2D and utilize its outputs to provide extensive pre-trained domain knowledge, ensuring robust and accurate feature learning in the student model.
\par

We employ a proxy dataset $D^{P}$ to conduct knowledge distillation and use Kullback-Leibler (KL) divergence to measure and minimize the discrepancy between the predictions of SAM-Med2D (teacher $\theta_{T}$) and SM-lite (student $\theta_{S}$). The KL divergence provides a way to quantify the difference in the predicted probability distributions of the two models, focusing on aligning the student model's predictions with those of the teacher. This alignment is achieved by optimizing the student model to minimize the KL divergence between the two sets of predictions. The process can be formalized as follows:
\begin{equation}
        \underset{\theta_{S}}{min} \underset{x \sim D^{P}}{\mathbb{E}} \left[ D_{KL} \left[ \sigma \left(g(x;\theta_{T})\right) \parallel \sigma \left(g(x;\theta_{S}) \right) \right] \right].
\end{equation}
Here, $g(\cdot)$ is the logits output, and $\sigma(\cdot)$ is the non-linear activation.
\par

In addition to the distillation loss, we also incorporate the ground truth segmentation masks into the loss function. This dual-loss approach, combining distillation loss with traditional supervised learning loss, ensures that SM-lite not only learns to mimic the teacher model's predictions but also adheres to the groundtruth. The complete loss function can be expressed as a weighted sum of the distillation loss and the segmentation loss (dice loss) against the groundtruth:
\begin{equation}
\mathcal{L} = \alpha \mathcal{L}_{\text{distill}} + (1 - \alpha) \mathcal{L}_{\text{segment}}(y, \sigma \left(g(x; \theta_{S}) \right) ).
\end{equation}
Here, $\mathcal{L}_{\text{distill}}$ is the distillation loss, $\mathcal{L}_{\text{segment}}$ is the segmentation loss, and $\alpha$ is a hyperparameter that balances the two types of loss.

\subsection{Initialization Strategies}
We consider three initialization strategies in our study: random initialization, pre-training initialization, and FM-instructed initialization.

\subsubsection{Random Initialization}
Random initialization in FL is characterized by starting the model training with weights that are randomly assigned. This strategy, devoid of any preliminary knowledge about the data or task, is fundamental in its approach and execution.
Its inherent randomness can lead to variability in performance, especially in non-IID data scenarios, often resulting in extended training times and potential convergence issues.
We use random initialization to serve as a baseline for comparison against more informed initialization strategies.

\subsubsection{Pre-training Initialization}
Pre-training initialization involves using a proxy dataset $D^P$ to pre-train the model before it is incorporated into FL. This process begins by selecting a representative dataset that closely aligns with the target task but is not part of the distributed datasets in the FL network. The model is initially trained on the proxy dataset to acquire a basic understanding of the task, which includes learning general features and patterns relevant to medical image segmentation. This foundational training phase aims to provide the model with a head start when it enters the FL cycle, potentially leading to quicker convergence and improved performance compared to models that begin with random weights.

\subsubsection{FM-instructed Initialization}
FM-instructed initialization represents a more advanced strategy, leveraging the vast pre-trained knowledge encapsulated in an FM like SAM-Med2D. The process involves distilling the knowledge from SAM-Med2D into a smaller model SM-lite as described in Section 2.2, which is then used as the starting point in FL. This distillation process is crucial; it involves training SM-lite on a proxy dataset $D^P$, ensuring that it learns to mimic the high-level feature representations and predictions of the SAM-Med2D model. The resulting SM-lite model embodies the extensive learning and generalization capabilities of SAM-Med2D but in a more compact and FL-friendly form. This strategy is potentially beneficial for dealing with the non-IID data challenge in FL, as it provides the model with a robust and comprehensive understanding of medical image segmentation tasks right from the beginning.


\section{Expriments and Results}
\subsection{Dataset and Implementation Details}
We utilize the COVID-19 chest X-ray dataset\footnote{https://www.v7labs.com/open-datasets/covid-19-chest-x-ray-dataset}, which contains 6,504 images of AP/PA chest X-ray annotated with pixel-level polygonal lung segmentations. 
All images are resized to 256$\times$256. Data augmentation, including rotation, flipping, and adjustments in brightness and contrast, are employed to enhance model generalizability \cite{li2022explainable}. 
The dataset is partitioned into three distinct segments: 1,504 images form the proxy dataset $D^P$, 500 images are reserved for testing, and the remaining 4,500 images are distributed across 3 clients.
To simulate the non-IID conditions, we design age skew and quantity skew.
Age skew refers to the imbalance in the age distribution of training data across 3 clients, while quantity skew refers to the imbalance in the number of training data across 3 clients.
\par

In the knowledge distillation process, the bounding box of groundtruth is used as the box prompt for SAM-Med2D, and $\alpha$ is set to 0.6 to balance the loss function.
Our FL model is based on the U-Net. Training involves 100 communication rounds, with each round consisting of 5 epochs. We use an Adam optimizer with an initial learning rate of 1e-4.
We also compare FedAvg employing various initialization strategies against centralized and standalone learning to assess their performance gap.
We use Dice score to evaluate the segmentation performance.
All the experiments are implemented in PyTorch and trained on 4 NVIDIA RTX 3090 GPUs.

\begin{table}[h]
    \centering
    \caption{The Segmentation performance of various initialization strategies for FedAvg (age skew).}
    \resizebox{0.35\textwidth}{!}{%
    \begin{tabular}{lc|cc}
    \toprule
    Initialization & Federated & Centralized & Standalone \\
    \midrule
    Random         & 74.67       & 89.41     & 67.20 \\
    Pre-training   & 83.92       & -         & - \\
    FM-instructed  & 85.38       & -         & - \\
    \bottomrule
    \end{tabular}
    }
\end{table}

\begin{figure}[!th]
    \centering
    \includegraphics[width=0.37\textwidth]{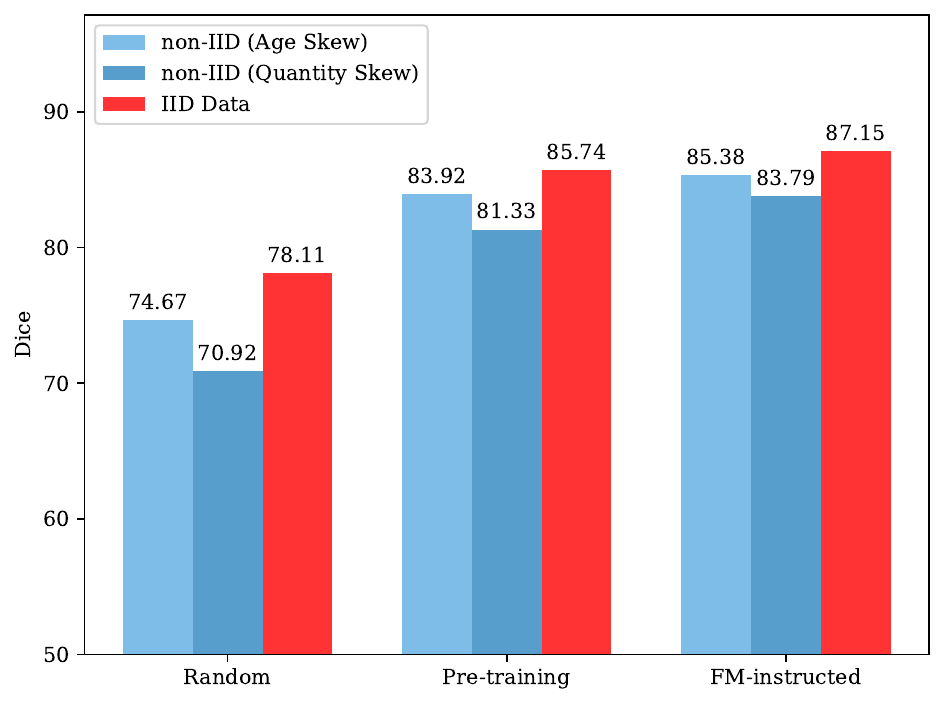}
    \caption{The segmentation performance of various initialization strategies for FedAvg trained on IID and non-IID data.}
\end{figure}

\begin{figure}[!th]
    \centering
    \includegraphics[width=0.37\textwidth]{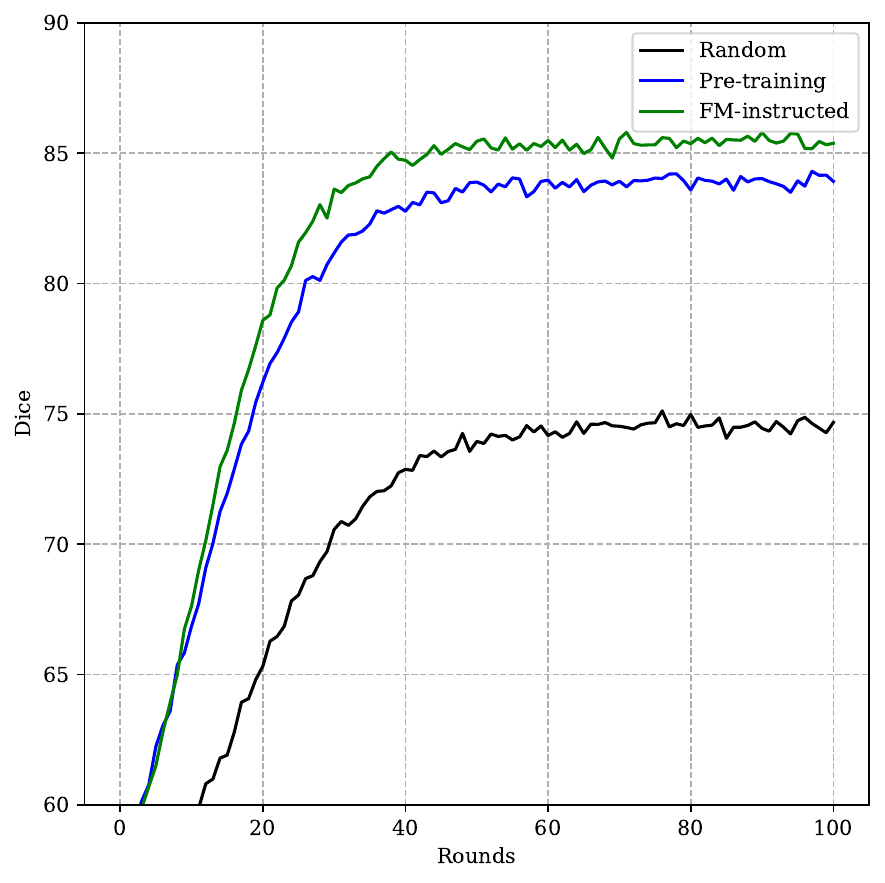}
    \caption{The training dynamics of various initialization strategies for FedAvg (age skew).}
\end{figure}

\subsection{Results: From Random to Pre-training and FM-instructed Initailization}

\noindent
\textbf{Reduce the performance gap between FL and centralized learning.}
The comparison of segmentation performance under different initialization strategies reveals that both Pre-training and FM-instructed initialization significantly reduce the performance gap between federated and centralized learning. As shown in Table 1, while the random initialization lags substantially behind the centralized approach with a Dice score of 74.67 in federated settings, both Pre-training and FM-instructed initialization markedly narrow this gap, achieving Dice scores of 83.92 and 85.38, respectively.  The latter score approaches the centralized learning of 89.41 more closely, suggesting that the FM-instructed initialization can further enhance the performance of the model.

\noindent
\textbf{Narrow the performance gap between non-IID and IID.}
We compare FedAvg employing various initialization strategies under IID and non-IID (age skew and quantity skew) data splits.
Figure 2 indicates that both Pre-training and FM-instructed initialization demonstrate potential capabilities to mitigate the non-IID issue. 
When comparing non-IID and IID scenarios, Pre-training and FM-instructed initialization show marginal decreases, unlike the random initialization which exhibits a clear performance drop, emphasizing the robustness of Pre-training and FM-instructed initialization against data heterogeneity.
Specifically, in non-IID (age skew) conditions, the decrement in performance is substantially less pronounced for FM-instructed initialization compared to Pre-training initialization, suggesting that the extensive pre-trained knowledge embedded within FM offers an advantage in managing data heterogeneity.

\noindent
\textbf{Faster convergence to better performance.}
Figure 3 presents the training dynamics of various initialization strategies, with a particular emphasis on the accelerated convergence rates facilitated by pre-training and, more notably, FM-instructed initializations. 
It indicates a steep curve for Pre-training and FM-instructed initialization, reaching higher Dice scores in fewer rounds (less than 40 rounds) compared to random initialization strategy (more than 40 rounds). This rapid convergence verifies the efficiency of more informed initialization strategies and further suggests that the FM-instructed initialization enhances the federated model's ability to start with a strong pre-trained knowledge base, leading to quickly achieving optimal performance.

\section{Conclusion}
Our study focuses on the impact of FM-instructed initialization in FL.
We find that FM-instructed initialization can serve as a good teacher to help address the suboptimal performance issue in FL.
This study represents a pioneering effort to combine the strengths of FM with FL, potentially setting a new standard for model initialization in FL settings.

\section{Compliance with ethical standards}
\label{sec:ethics}

This research study was conducted retrospectively using the COVID-19 Chest X-ray dataset made available in open access. Ethical approval was not required as confirmed by the license attached with the open access dataset.

\section{Acknowledgments}
\label{sec:acknowledgments}

This study was supported in part by the ERC IMI (101005122), the H2020 (952172), the MRC (MC/PC/21013), the Royal Society (IEC\textbackslash NSFC\textbackslash211235), the NVIDIA Academic Hardware Grant Program, the SABER project supported by Boehringer Ingelheim Ltd, Wellcome Leap Dynamic Resilience, and the UKRI Future Leaders Fellowship (MR/V023799/1).

\bibliographystyle{IEEEbib}
\bibliography{refs}

\end{document}